%% file: template.tex
\documentclass{article}

\usepackage{arxiv}

\usepackage[utf8]{inputenc} 
\usepackage[T1]{fontenc}    
\usepackage{hyperref}       
\usepackage{url}            
\usepackage{booktabs}       
\usepackage{amsfonts}       
\usepackage{nicefrac}       
\usepackage{microtype}      
\usepackage{lipsum}		
\usepackage{graphicx}
\usepackage{natbib}
\usepackage{doi}
\usepackage{xcolor}         
\usepackage{bm}
\usepackage{amsmath}
\usepackage{amsthm}
\usepackage{mathtools}
\usepackage{wrapfig}
\usepackage{bm}
\usepackage{multirow}
\usepackage{amsfonts}
\usepackage{thm-restate}
\usepackage{tabularx} 
\usepackage{tablefootnote}

\usepackage{colortbl}
\usepackage{amssymb}

\usepackage{algorithm}
\usepackage{float}
\usepackage{algorithmic}
\usepackage{caption}
\usepackage{xspace}

\usepackage{algorithm} 
\usepackage{algorithmic}
\hypersetup{colorlinks, linkcolor=blue, citecolor=blue, urlcolor=magenta, linktocpage, plainpages=false}
\graphicspath{{./figures/}}
\usepackage{multirow}

\definecolor{lightgrey}{HTML}{dcdbdb}
\definecolor{lightblue}{HTML}{E8F0FE}
\definecolor{lightblue}{HTML}{E8F0FE}
\definecolor{gray}{HTML}{9aa0a6}
\definecolor{lightpink}{HTML}{F48FB1}
\definecolor{lightred}{HTML}{FFCBC9}
\definecolor{lightcyan}{HTML}{80DEEA}
\definecolor{darkgreen}{rgb}{0.0, 0.5, 0.0}

\usepackage[skins]{tcolorbox} 
\tcbuselibrary{breakable} 
\newtcolorbox[auto counter, number within=section, list type=subsubsection, list inside=toc]{sectionbox}[2][]{
colback=white!98!gray, colframe=black, 
colbacktitle=white!90!gray, coltitle=black, 
fonttitle=\bfseries,
title={#2}, 
list entry={Comment \thetcbcounter\quad}
}
\usepackage{lipsum}
\usepackage{soul}

\definecolor{greengrey}{rgb}{0.0, 0.3, 0.0} 
\colorlet{greengreywithhint}{greengrey!90!gray} 

\title{\raisebox{-0.35cm}{\includegraphics[width=1.2cm]{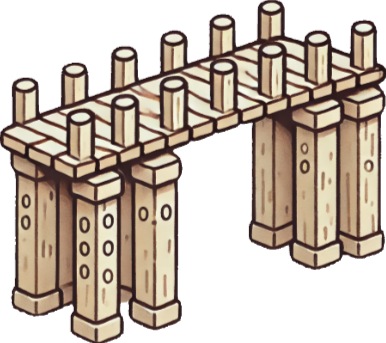}} Bridge-Coder: Unlocking LLMs' Potential \\to Overcome Language Gaps in Low-Resource Code}

\usepackage{multirow}
\usepackage{colortbl}
\usepackage{graphicx}
\usepackage{makecell} 

\title{\raisebox{-0.35cm}{\includegraphics[width=1.2cm]{figures/icon.jpg}} Bridge-Coder: Unlocking LLMs' Potential \\to Overcome Language Gaps in Low-Resource Code}

\author{\textbf{Jipeng Zhang}$^1$\footnotemark[1]\protect\phantom{\footnotesize 1}, \textbf{Jianshu Zhang}$^{1}$$^{*}$, \textbf{Yuanzhe Li}$^{3}$\thanks{\, Equal Contribution. Code are available at the following links: \url{https://github.com/2003pro/bridgecoder}.}\protect\phantom{\footnotesize 1}, \textbf{Renjie Pi}$^1$, \textbf{Rui Pan}$^2$, \textbf{Runtao Liu}$^1$, \textbf{Ziqiang Zheng}$^1$, \\
\textbf{Tong Zhang}$^2$
\\
  $^1$The Hong Kong University of Science and Technology\\
 $^2$University of Illinois Urbana-Champaign \\
 $^3$Southern University of Science and Technology \\
\texttt{\{jzhanggr,rpi,rliuay,zzhengaw\}@ust.hk,} \\ \texttt{jianshu.zhang@whu.edu.cn,} \texttt{12113016@mail.sustech.edu.cn,} \\ \texttt{ruip4@illinois.edu,}
\texttt{tongzhang@tongzhang-ml.org}
}

\renewcommand{\shorttitle}{}

\hypersetup{
pdftitle={A template for the arxiv style},
pdfsubject={q-bio.NC, q-bio.QM},
pdfauthor={David S.~Hippocampus, Elias D.~Striatum},
pdfkeywords={First keyword, Second keyword, More},
}

\begin{document}
\maketitle
\begin{abstract}
Large Language Models (LLMs) demonstrate strong proficiency in generating code for high-resource programming languages (HRPLs) like Python but struggle significantly with low-resource programming languages (LRPLs) such as Racket or D. 
This performance gap deepens the digital divide, preventing developers using LRPLs from benefiting equally from LLM advancements and reinforcing disparities in innovation within underrepresented programming communities.
While generating additional training data for LRPLs is promising, it faces two key challenges: manual annotation is labor-intensive and costly, and LLM-generated LRPL code is often of subpar quality.
The underlying cause of this issue is the gap between natural language to programming language gap (NL-PL Gap), which is especially pronounced in LRPLs due to limited aligned data.
In this work, we introduce a novel approach called \textbf{Bridge-Coder}, which leverages LLMs' intrinsic capabilities to enhance the performance on LRPLs. Our method consists of two key stages. \textit{Bridge Generation}, where we create high-quality dataset by utilizing LLMs' general knowledge understanding, proficiency in HRPLs, and in-context learning abilities. Then, we apply the \textit{Bridged Alignment}, which progressively improves the alignment between NL instructions and LRPLs.
Experimental results across multiple LRPLs show that Bridge-Coder significantly enhances model performance, demonstrating the effectiveness and generalization of our approach. Furthermore, we offer a detailed analysis of the key components of our method, providing valuable insights for future work aimed at addressing the challenges associated with LRPLs.
\end{abstract}

\section{Introduction}
\begin{figure}[!t]
    \begin{center}
    \includegraphics[width=0.45\columnwidth]{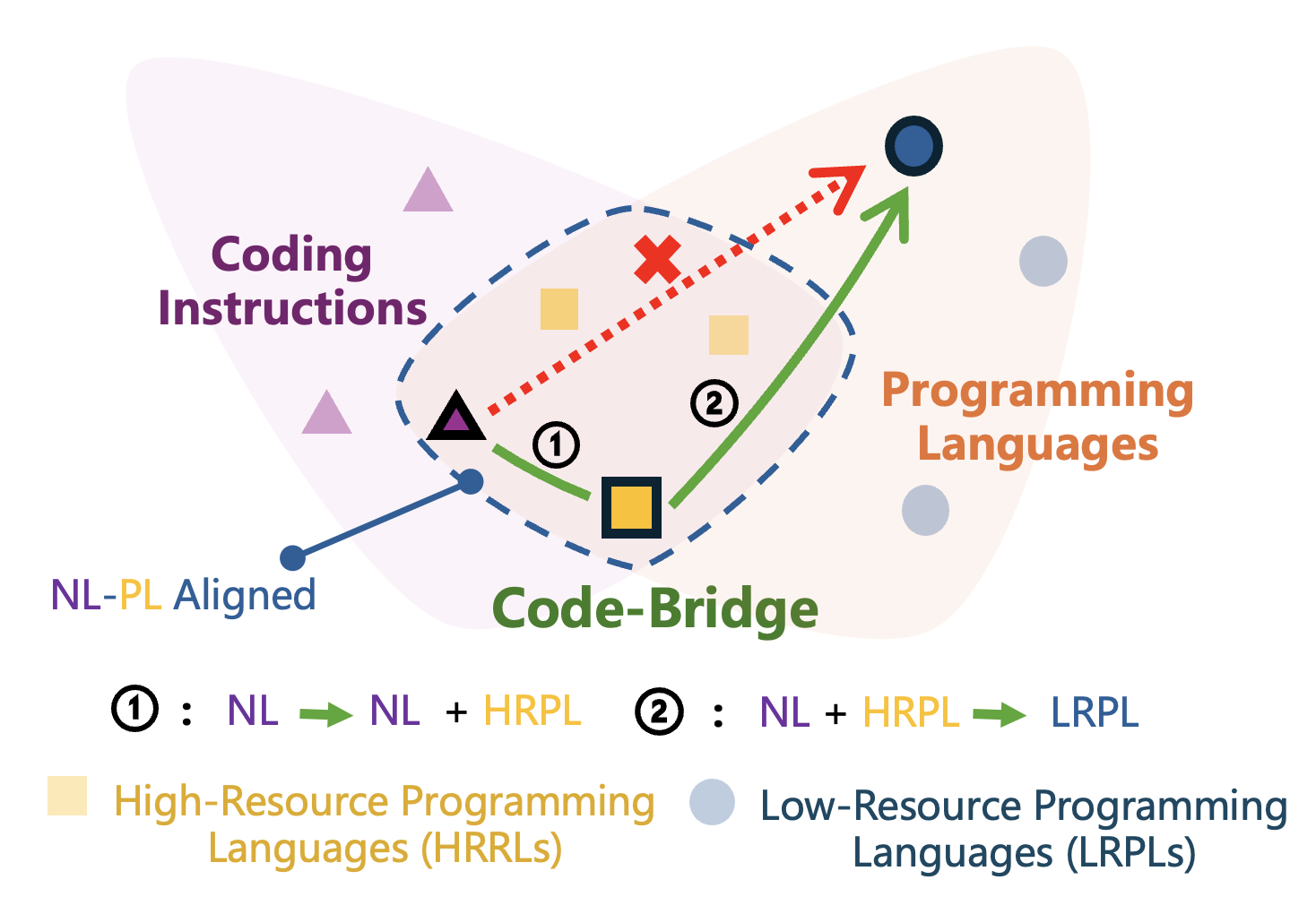}
    \caption{An illustration of how \textcolor{darkgreen}{code-bridge} helps solve tasks in low-resource programming languages (LRPLs). \textcolor{red}{Directly generating responses} leads to suboptimal results, as models struggle to follow instructions accurately in LRPLs due to limited training data. In contrast, using \textcolor{darkgreen}{code-bridge} can improve performance by first generating code and comments in high-resource programming languages (HRPLs), using this output as a bridge to guide the model toward producing more accurate and coherent responses in LRPLs.}
    \label{fig:intro}
    \end{center}
    \vskip -1 em
\end{figure}

Large Language Models (LLMs) have shown remarkable capabilities~\citep{chen2021humaneval} in assisting with coding tasks such as code generation~\citep{austin2021program}, debugging~\citep{zhong2024ldebug,xia2024fuzz4all}, code explanation~\citep{Nam2023UsingAL}. With the introduction of tools like GitHub Copilot~\citep{copilot,whisper-copilot}, models like OpenAI Codex~\citep{chen2021humaneval} have greatly enhanced the efficiency of developers by automating repetitive tasks, providing real-time code suggestions, and offering in-depth explanations of code functionality.

However, when it comes to low-resource programming languages (LRPLs), the instruction-following abilities of LLMs are significantly diminished~\citep{cassano2022multiple}. This limits LLMs' ability to effectively support developers working with LRPLs~\citep{CodeGeeX,orlanski2023measuring,bansal2024llm}, preventing these developers from fully benefiting from the advanced capabilities that LLMs provide for High-Resource Programming Languages (HRPLs). This uneven distribution of benefits exacerbates digital inequality, further widening the gap between developers in different programming ecosystems.

While some research has been conducted on Low-Resource Natural Languages (LRNLs)~\citep{xue2020mt5,lample2019cross, Huang2019UnicoderAU, Conneau2019UnsupervisedCR, Hu2020XTREMEAM}, LRPLs remain relatively under-explored. The challenges that LLMs face with LRPLs are primarily stemming from data imbalances in representation~\citep{hangya-etal-2022-improving,Wang2020ExtendingMB}. However, programming languages (PLs) present an additional complexity: unlike natural language (NL) training data, which is purely in NL, training data for programming languages often includes both PL (the code) and NL (such as instructions or comments). This interplay between NL and PL in programming languages makes addressing the LRPL gap more intricate.

This introduces what we refer to as the NL-PL Gap—the inherent disconnect caused by the need to align natural languages with programming languages. Consequently, even though LLMs indeed possess a certain level of proficiency in LRPLs, they often struggle to follow NL-based instructions due to the added complexity of the NL-PL Gap. As illustrated in Figure \ref{fig:intro}, directly generating responses in LRPLs frequently results in low-quality outputs because of this gap. However, by first generating combined NL-PL results in HRPLs and then using those as a reference to generate LRPL outputs, the gap can be effectively bridged, leading to improved performance.

In light of this, we propose \textbf{Bridge-Coder}, which aims to mitigate the gap between HRPLs and LRPLs in LLMs. Specifically, Bridge-Coder consists of two key phases: 
1) \textit{Bridge-Assisted Generation}, where we start by leveraging the LLMs' general reasoning abilities for task screening, ensuring that the selected tasks are answerable within the LRPL context. Next, the LLM generates an answer in an HRPL, which includes both the code and NL-formatted comments explaining the solution. This combination of code and comments forms the code-bridge. The LLM then leverages its in-context learning ability to reference this code-bridge when responding to instructions in LRPLs, allowing it to generate more accurate and coherent results.
2) \textit{Bridged Alignment} begins with assist alignment, where we help the LLM gradually bridge the NL-PL Gap by providing intermediate guidance, ensuring the model doesn’t face too large a leap at once. Following this, the focus shifts to direct alignment, enhancing the model's ability to independently respond to NL instructions in LRPLs. 


Extensive experimental results across various LRPLs demonstrate significant improvements in the LLM's performance, particularly in its ability to follow instructions. Additionally, we conducted comprehensive experiments to validate the technical choices within our Bridge-Coder framework. These experiments provide valuable insights into the underexplored area of low-resource programming languages, revealing opportunities for further advancement.

In summary, our contributions are:
\begin{itemize}
    \item We identify the NL-PL Gap as the primary factor behind LLMs' poor performance in LRPLs. This gap emerges due to programming language datasets containing both PL and NL components (e.g., instructions, comments), complicating the alignment between NL instructions and PL outputs.
    \item We introduce Bridge-Coder, a two-stage method designed to enhance performance in LRPLs, which first leverages LLMs' potential to generate high-quality data, then aligns the models with the NL-PL Gap through assist and direct alignment steps.
    \item Through extensive experiments across various LRPLs, we demonstrate the effectiveness and generalization of Bridge-Coder. Moreover, we provide valuable insights that can guide future research in the underexplored field of low-resource programming languages.
\end{itemize}

\section{Related Work}
\subsection{Code LLMs}

\paragraph{Foundation Code Models.} Training on code samples with billions of tokens using hundreds of high-performance GPUs, decoder-only code foundation LLMs have been proven to have strong code generation ability across various tasks. Specifically, Codex~\citep{codex} is OpenAI's earliest domain-specific LLM for coding and is believed to support the Copilot service, which helps with automatic code completion across different IDEs~\citep{copilot}. Additionally, the open-source community has developed a series of code LLMs, such as InCoder~\citep{incoder} and CodeT5~\citep{codet5}, to further support the development of stronger or domain-specific code assistants. More precisely, Deepseek-coder~\citep{deepseek-coder} family models and StarCoder~\citep{li2023starcoder} family models trained their model parameters from scratch with trillions of tokens scraped from web pages related to code. Code-Llama~\citep{roziere2023codellama} and Code-Qwen~\citep{hui2024codeqwen2} family models perform post-training from general-purpose models with code-related datasets to achieve high-performance code foundation models.

\paragraph{Finetuning Code Models.} Besides developing code foundation models, researchers often finetune these code foundation models for specific applications. Maigcoder~\citep{wei2023magicoder} utilizes open-source code snippets to create instruction datasets for further improving code LLMs' instruction-following abilities. Wizard-Coder~\citep{luo2023wizardcoder} and WavCoder~\citep{yu2023wavecoder} use evol-instruct~\citep{xu2023wizardlm} to extract effective instruction-code pairs from proprietary LLMs through few-shot prompting and self-improvement. OctoCoder~\citep{muennighoff2023octopack_OctoCoder} uses Git commits and code changes to generate instruction-following data and enhance the model's coding ability. Besides these works, there exist several works~\citep{paul2024ircoder,sun2024unicoder} focusing on using intermediate representation like from LLVM to improve Code LLMs. \citet{cassano2024knowledge} proposed to translate high resource PLs to low resource PLs with the help of compiler. 

In contrast to works that focus solely on improving code completion abilities in PLs~\citep{cassano2024knowledge}, our research aims to enhance code LLMs' instruction-following capability for LRPLs. We achieve this by improving the alignment between NL and PL during the fine-tuning stage.

\subsection{LLMs' Intrinsic Capabilities}
Large Language Models (LLMs) possess several intrinsic capabilities that are a result of both the extensive training they undergo and the fine-tuning they receive on specialized data. One of their core strengths is general knowledge reasoning~\citep{liang2022helm}, which arises from the vast amount of diverse data they are trained on~\citep{touvron2023llama,DBLP:journals/corr/llama2,deepseek-coder}. This general reasoning ability enables LLMs to provide accurate responses to a wide range of tasks across different domains, even without task-specific fine-tuning.

In addition to general reasoning, LLMs can develop proficient knowledge when fine-tuned on domain-specific datasets. This allows them to excel in specialized areas, such as legal~\citep{colombo2024saullm_lawllm}, medical~\citep{singhal2023medpalm,diao2023lmflow}, or technical fields~\citep{taylor2022galactica}, where deep and nuanced understanding is required. Fine-tuning equips LLMs with the ability to handle more complex and precise tasks that go beyond their general knowledge base.

Another powerful capability of LLMs is In-Context Learning (ICL)~\citep{GPT3}. ICL enables models to generate more accurate responses by learning from the context provided in the input, without the need for further training. As a training-free approach, ICL is highly flexible and can be applied in various ways, including data generation~\citep{wang2022self, ye2022zerogenefficientzeroshotlearning,
gao2023selfguidednoisefreedatageneration, wang2024theoremllamatransforminggeneralpurposellms, pi2024imagetextualizationautomaticframework, gao2023gllavasolvinggeometricproblem, pi2024strengtheningmultimodallargelanguage},  personalized conversations~\citep{pi2024personalized}, where the model adapts to user preferences; and task-specific guidance, where context helps refine and improve response accuracy. ICL's versatility makes it a valuable tool for enhancing performance across different applications.

\begin{table}[t]
\centering
\setlength{\extrarowheight}{2pt} 
\begin{tabular}{lcccc}
\toprule
\multirow{2}{*}{Language} & \multicolumn{2}{c}{StarCoder} & \multirow{2}{*}{GitHub (\%)} & \multirow{2}{*}{TIOBE (\%)} \\ 
\cline{2-3}
 & Num. files & Percentage & & \\
\hline 
Bash & - & - & - & 43 \\
C++ & 6,377,914 & 6.379 & 7.0 & 4 \\
C\# & 10,839,399 & 5.823 & 3.1 & 5 \\
D & - & - & - & 35 \\
Go &4,730,461  & 3.101 & 7.9 & 12 \\
Java & 20,151,565 & 11.336 & 13.1 & 3 \\
JavaScript & 21,108,587 & 8.437 & 14.3 & 7 \\
Python & 12,962,249 & 7.875 & - & 1 \\
R & 39,194 & 0.039 & 0.05 & 19 \\
Racket & 4,201 & 0.004 & - & - \\
Rust & 1,386,585 & 1.188 & 1.1 & 22 \\
Ruby &3,405,374 & 0.888 & 6.2 & 15 \\

\bottomrule
\end{tabular}%
\vspace{0.4cm}
\caption{Programming Language Statistics: The StarCoder parts are based on data from their report~\citep{li2023starcoder}. The last two columns are derived from GitHut 2.0 and the 2022 TIOBE Programming Community Index, as referenced in the MultiPLE benchmark paper~\citep{cassano2022multiple}.}
\label{tab:language_stats}
\end{table}

\section{NL-PL Gap}

The NL-PL Gap refers to the disparity that arises when LLMs are tasked with following natural language (NL) instructions in programming language (PL). This gap is particularly pronounced in low-resource programming languages (LRPLs). The NL-PL gap stems from the following key factors:

\begin{figure*}[t]
\centering
\includegraphics[width=0.97\textwidth]{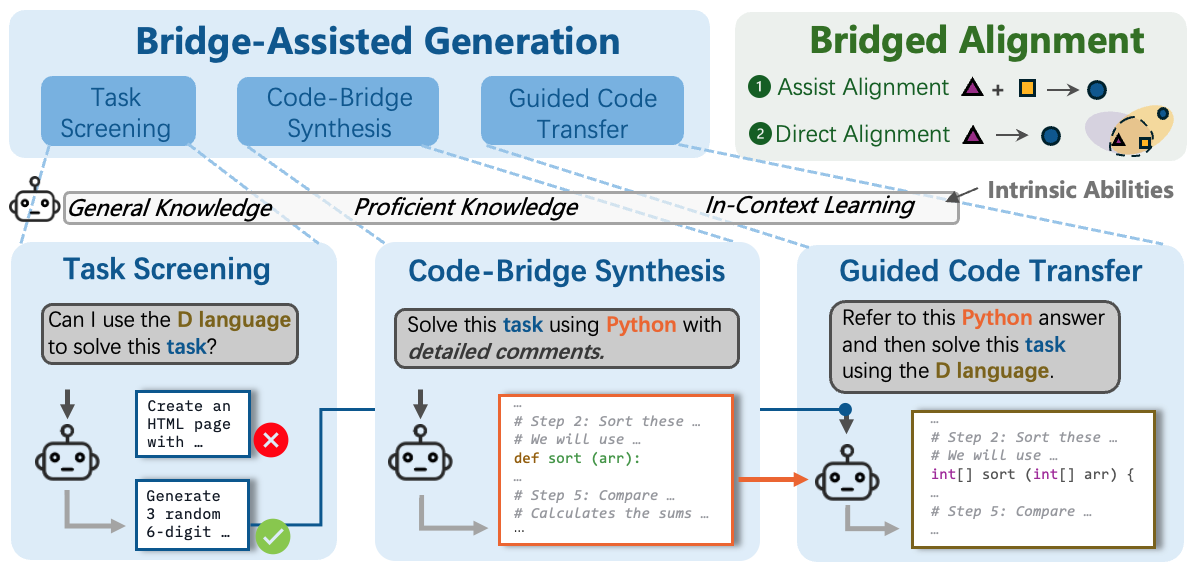} 
\caption{An illustration of \textbf{Bridge-Coder}. In \textit{Bridge-Assisted Generation}, the LLM first identifies tasks suitable for the target low-resource programming language (LRPL). Then, it generates a code-bridge in a high-resource programming language (HRPL), combining both code and comments to explain the solution. This code-bridge is then used to help bridge the NL-PL gap in LRPLs. In \textit{Bridged Alignment}, the model is first guided by the code-bridge to assist in aligning the NL-PL gap, and later progresses to generating responses directly from natural language instructions without the code-bridge.}
\label{fig:pipeline}
\end{figure*}

\paragraph{Data Imbalance.} The statistics in Table \ref{tab:language_stats} highlights the stark data imbalance between high-resource and low-resource programming languages. Languages like JavaScript, Python, and Java have millions of files in the StarCoder dataset, providing LLMs with extensive NL-PL aligned data. In contrast, low-resource languages such as R, Racket, and D are vastly underrepresented, with only a fraction of the data available. This disparity limits the model's ability to learn effective mappings from NL to PL in LRPLs, significantly contributing to the NL-PL Gap. The GitHub and TIOBE indices further reflect this imbalance, reinforcing the challenges faced by LLMs when generating code for underrepresented languages.

\paragraph{Complexity of Mapping NL to PL.} Unlike purely natural language tasks, coding tasks require models to first understand NL instructions and then generate executable PL code, often together with NL comments to explain the code. In HRPLs, models excel due to the abundance of NL-PL aligned data. However, for LRPLs, this mapping is more difficult due to limited data. As shown in our experiments in Section \ref{exp: main}, directly generating code for LRPLs leads to lower-quality outputs, whereas using a code-bridge as a transitional step improves code quality by mitigating the NL-PL gap.

\section{Bridge-Coder}
In this section, we present the details of \textbf{Bridge-Coder}, including two phases: Bridge-Assisted Generation (Section \ref{sec: bridge-assisted generation}) and Bridged Alignment (Section \ref{sec: bridged-training}). The key idea of Bridge-Assisted Generation is fully leveraging LLMs' intrinsic capabilities to generate instruction following data for low-resouce programming languages (LRPLs). Afterward, Bridged Alignment gradually helps the model overcome the NL-PL Gap, improving its performance on LRPLs. An illustration of Bridge-Coder is shown in Figure \ref{fig:pipeline}.

\subsection{Bridge-Assisted Generation}
\label{sec: bridge-assisted generation}
This section introduces a novel approach for generating training data for LRPLs. We first leverage the LLM's general knowledge reasoning abilities to identify the task set \(\mathcal{T}\) that can be effectively solved using the target LRPL, denoted as \(PL_{\text{tar}}\). Next, we utilize the LLM's strong understanding and generation capabilities in a HRPL to generate the code-bridge, denoted as \(PL_{\text{bdg}}\). Finally, by using the LLM’s in-context learning (ICL) abilities, we rephrase the task  \(\mathcal{T}\) with the help of \(PL_{\text{bdg}}\), which enables the LLM to generate the desired response in the target LRPL \(PL_{\text{tar}}\).

\subsubsection{Task Screening}
Existing code instruction datasets often include general-purpose tasks, while others can only be solved with specific programming languages. If unsuitable tasks are not filtered out, the model may fail to respond when the task is unanswerable in the target language. Even more concerning, several studies~\citep{ spiess2024quality, shum2024first} have highlighted that, due to mis-calibration, LLMs tend to confidently generate incorrect answers in such cases. This issue further emphasizes the importance of task screening to prevent such errors and improve response quality. 

We observe that while current LLMs struggle with LRPL code generation, they perform much better in classification tasks that simply judge whether a task can be solved using a specific LRPL. This is because classification tasks, unlike code generation, do not require the LLM to bridge the NL-PL Gap. Instead, the model can rely on its general reasoning abilities to provide a straightforward `Yes' or `No'. answer. Additionally, as the model's accuracy improves, we enhance this process by requiring the LLM to provide logical explanations for its judgments, further validating its decision-making process.
We validate the importance of this screening step with experimental evidence in Section \ref{analysis: task screening}, showing its critical role in improving output quality.

\subsubsection{Code-Bridge Synthesis}
When LLMs answer task \(\mathcal{T}\), they first need to comprehend the natural language (NL) instructions and then generate a response in the programming language (PL), which often includes adding NL comments to the code. This makes NL-PL alignment in the training data crucial. In high-resource programming languages (HRPLs), the abundance of NL-PL aligned data in the training sets allows LLMs to perform effectively.

Here, we leverage the existing capabilities of LLMs in HRPLs to follow the NL instruction. Furthermore, we also ask LLMs to include comments explaining the key steps and the thought process behind the solution. In this way, we create the code-bridge \(PL_{\text{bdg}}\), which combines both NL (i.e., comments) and PL (i.e., code). This serves as a reinterpretation of the NL instruction from the perspective of PL logic, transforming what might seem like simple instructions in NL into a more explicit and detailed process in PL.
Programming languages often require a step-by-step breakdown and precise logic that natural language tends to abstract away, making \(PL_{\text{bdg}}\) an essential way for bridging this gap.
This approach ensures that even with limited NL-PL aligned data in LRPLs, LLMs can still effectively generate correct and coherent solutions by leveraging the detailed structure and reasoning provided by the code-bridge.


\subsubsection{Guided Code Transfer}
For LRPLs, which are underrepresented in training data, the NL-PL Gap presents a major challenge. Although LLMs possess some code generation capabilities, the lack of well-aligned data between NL instructions and PL reasoning leads to suboptimal solutions, making it difficult to generate accurate responses to NL instructions in LRPLs.

To overcome this, we utilize the previously generated code-bridge to mitigate the NL-PL gap. During this step, when generating responses in LRPLs, the code-bridge is appended to the instruction as additional context. By harnessing the in-context learning (ICL) capabilities of LLMs, this approach allows the model to reference the PL logic in the code-bridge, guiding it when responding to NL instructions. This significantly improves the quality of the model's output in LRPLs.

This process is analogous to a non-native English speaker first drafting their thoughts in their native language and then translating them into English. The code-bridge acts as a ``draft" in PL, enabling the LLM to better interpret NL instructions and produce more accurate answers in LRPLs.

\subsection{Bridged Training}
\label{sec: bridged-training}

We draw inspiration from the concept of curriculum learning~\citep{bengio2009curriculum} and apply it to the learning of LRPLs. To effectively bridge the NL-PL gap and improve LLM performance in low-resource programming languages, we divide the training process into two stages. 

\paragraph{Assist Alignment.} In the first stage, the primary goal is to assist the LLM in bridging the NL-PL gap by providing additional support through the code-bridge.
The input includes the instruction of task $\mathcal{T}$, along with the code-bridge in the high-resource programming language $PL_{\text{bdg}}$, which serves as a guide. The LLM uses this reference to assist in generating the target response in the low-resource programming language $PL_{\text{tar}}$. 
The loss function can be formalized as:

\vspace{-1.5em}
$$
    \mathcal{L}_{\text{assist}} = - \sum_{t=1}^{T} \log P(y_t^{PL_{\text{tar}}} \mid y_{<t}^{PL_{\text{tar}}}, \mathcal{T}, PL_{\text{bdg}})
$$

\paragraph{Direct Alignment.} In the second stage, the focus shifts to helping the LLM adapt to real-world scenarios by asking it to directly follow NL instructions without any assistance from the code-bridge. This approach ensures the model becomes more capable of solving tasks independently, as it would in practical applications where such assistance is not available. The training loss for this phase is calculated as:
\vspace{-0.7em}
\[
\mathcal{L}_{\text{direct}} = - \sum_{t=1}^{T} \log P(y_t^{PL_{\text{tar}}} \mid y_{<t}^{PL_{\text{tar}}}, \mathcal{T})
\]

This two-step process facilitates a smooth and effective learning progression, moving from guided learning with assistance to independent problem-solving in LRPLs, as validated in the subsequent experiments in Section \ref{sec: training exp}, highlighting the benefits of this approach.

\section{Evaluation Settings}

\subsection{LRPLs and Benchmarks}
\paragraph{LRPLs.} To fully validate the generalization ability of our method, we selected four low-resource programming languages: R, D, Racket, and Bash. These languages cover a broad range of functionalities, including statistical computing, systems programming, language creation, and automation. Detailed descriptions of these languages can be found in the Appendix \ref{appendix: lrpl}.

\begin{table*}[t]
\centering
\setlength{\extrarowheight}{2pt} 
\begin{tabular}{c|cccc|c}
\Xhline{4\arrayrulewidth}  
\textbf{Models} & \textbf{R} & \textbf{D} & \textbf{Bash} & \textbf{Racket} & \textbf{Average}\\ 
\Xhline{3\arrayrulewidth}
\midrule
\multicolumn{6}{c}{\textbf{M-HumanEval}} \\
\midrule
DeepSeek-Coder-Base & 29.28 & 20.73 & 24.51 & 19.84 & 23.59\\
Magicoder-DS & 38.31 \textcolor{darkgreen}{+9.03} & 19.47 \textcolor{red}{-1.26} & 29.17 \textcolor{darkgreen}{+4.66} & 29.17 \textcolor{darkgreen}{+9.33} & 29.03 \textcolor{darkgreen}{+5.44}\\
Magicoder-S-DS & 40.63 \textcolor{darkgreen}{+11.35} & 24.60 \textcolor{darkgreen}{+3.87} & 33.06 \textcolor{darkgreen}{+8.55} & 30.50 \textcolor{darkgreen}{+10.66} & 32.20 \textcolor{darkgreen}{+8.61}\\
DeepSeek-Coder-DG & 37.56 \textcolor{darkgreen}{+8.28} & 27.76 \textcolor{darkgreen}{+7.03} & 37.26 \textcolor{darkgreen}{+12.75} & 30.95 \textcolor{darkgreen}{+11.11} & 33.38 \textcolor{darkgreen}{+9.79}\\
DeepSeek-Coder-BC & \textbf{49.11 \textcolor{darkgreen}{+19.83}} & \textbf{35.51 \textcolor{darkgreen}{+14.78}} & \textbf{42.99 \textcolor{darkgreen}{+18.48}} & \textbf{41.57 \textcolor{darkgreen}{+21.73}} & \textbf{42.30 \textcolor{darkgreen}{+18.71}}\\
\midrule
\multicolumn{6}{c}{\textbf{M-MBPP}} \\
\midrule
DeepSeek-Coder-Base & 38.23 & 30.83 & 28.67 & 32.48 & 32.55\\
Magicoder-DS & 41.13 \textcolor{darkgreen}{+2.90} & 32.49 \textcolor{darkgreen}{+1.66} & 28.52 \textcolor{red}{-0.15} & 37.75 \textcolor{darkgreen}{+5.27} & 34.97 \textcolor{darkgreen}{+2.42}\\
Magicoder-S-DS & 44.03 \textcolor{darkgreen}{+5.80} & 31.76 \textcolor{darkgreen}{+0.93} & 24.43 \textcolor{red}{-4.24} & 37.83 \textcolor{darkgreen}{+5.35} & 34.51 \textcolor{darkgreen}{+1.96}\\
DeepSeek-Coder-DG & 46.74 \textcolor{darkgreen}{+8.51} & 36.47 \textcolor{darkgreen}{+5.64} & 34.26 \textcolor{darkgreen}{+5.59} & 33.96 \textcolor{darkgreen}{+1.48} & 37.86 \textcolor{darkgreen}{+5.31}\\
DeepSeek-Coder-BC & \textbf{50.53 \textcolor{darkgreen}{+12.30}} & \textbf{43.51 \textcolor{darkgreen}{+12.68}} & \textbf{35.83 \textcolor{darkgreen}{+7.16}} & \textbf{43.57 \textcolor{darkgreen}{+11.09}} & \textbf{43.36 \textcolor{darkgreen}{+10.81}}\\
\Xhline{3\arrayrulewidth}  
\end{tabular}
\vspace{0.2cm}
\caption{Comparison of different models across M-HumanEval and M-MBPP (R, D, Bash, Racket). All models are based on DeepSeek-Coder-Base. DeepSeek-Coder-DG is obtained by fine-tuning DeepSeek-Coder-Base using data generated through direct generation, while DeepSeek-Coder-BC is derived from our Bridge-Coder framework. \textbf{Bold} values indicate the best performance. \textcolor{darkgreen}{$+$} and \textcolor{red}{$-$} represent the difference compared to DeepSeek-Coder-Base.}
\label{exp: main exp}
\end{table*}

\paragraph{Benchmarks.}
We utilized the adapted versions of two benchmarks, HumanEval~\citep{chen2021humaneval} and MBPP~\citep{MBPP}, which were originally designed for Python. These versions, provided by MultiPL-E~\citep{cassano2022multiple}, extend the original benchmarks by using lightweight compilers to translate them into multiple programming languages. We refer to these translated versions as M-HumanEval and M-MBPP, representing the multilingual counterparts of the original benchmarks.

\subsection{Models}

\paragraph{Baseline Models.} 
We selected three baseline models for comparison: DeepSeek-Coder-Base~\citep{deepseek-coder}, a foundational Code LLM on which our subsequent experiments are fine-tuned; Magicoder-DS~\citep{wei2023magicoder}, a widely used benchmark model for code generation, trained on the OSS-Instruct dataset and also built on the DeepSeek-Coder-Base; and MagicoderS-DS~\citep{wei2023magicoder}, an enhanced version of Magicoder-DS, trained on both OSS-Instruct and Evol-Instruct datasets~\citep{luo2023wizardcoder}, which offers superior performance across various coding benchmarks and is also based on the DeepSeek-Coder-Base.

\paragraph{Models for Generation.} 
The models used during the Bridged-Assisted Generation process include: 
Llama3-70B~\citep{dubey2024llama3}, which is the primary model we use. It can be considered both a code LLM and a general-purpose LLM, with strong performance across various tasks. 
Llama3-8B, which leans more towards being a general LLM with moderate code generation capabilities. 
StarCoder2-Instruct-15B~\citep{li2023starcoder}, a specialized code LLM, with strong capabilities in code-related tasks but limited general-purpose abilities.

\subsection{Comparison}
\paragraph{Data Generation.}
In our experiments, we employ two different approaches for generating data: DG (Direct Generation): In this approach, the model generates code directly from the natural language task without any intermediate steps. BC (Bridge-Coder): This approach introduces a code-bridge, which acts as an intermediary between the task and the final code generation. First, the model generates an intermediate code-bridge, which is then used to produce the final output.

\paragraph{Training Methods.}
We compare several training techniques, where our training data is represented as \(\{ \text{input}; \text{output} \}\). \(\mathcal{T}\) denotes the NL (natural language) task instruction, \(PL_{\text{bdg}}\) represents the High-Resource PL output that acts as a bridge, and \(PL_{\text{tar}}\) is the answer in the target low-resource programming language (LRPL). The \textit{Separate Alignment} method is represented as \(\{ \mathcal{T}; PL_{\text{bdg}} \} \cup \{ \mathcal{T}; PL_{\text{tar}} \}\). \textit{Direct Alignment} involves \(\{ \mathcal{T}; PL_{\text{tar}} \}\). \textit{Assist Alignment} combines \(\{ \mathcal{T}, PL_{\text{bdg}}; PL_{\text{tar}} \}\), while \textit{Bridged Alignment} begins with assist alignment and transitions to direct alignment.

\section{Experimental Results}

\subsection{Main Results}
\label{exp: main}
As shown in Table \ref{exp: main exp}, while the Magicoder series models~\citep{wei2023magicoder} demonstrate strong overall capabilities in code generation tasks for HRPLs, they struggle to consistently maintain this performance when applied to LRPLs.  For instance, Magicoder-DS and Magicoder-S-DS even perform worse than their baseline model DeepSeek-Coder-Base~\citep{deepseek-coder} in D and Bash. This can be attributed to the fact that their training data remains heavily focused on HRPLs, which limits their ability to generalize to LRPLs.
On the other hand, while DeepSeek-Coder-DG, which is fine-tuned on LRPL-specific data generated via the Direct Generation (DG) approach, shows some improvement over the baseline, the gains are minimal. In some cases, it even performs worse than the Magicoder models, which were trained primarily on HRPLs. This highlights that training with directly generated LRPL data is not the most effective strategy.

In contrast, applying the Bridge-Coder approach to DeepSeek-Coder-Base, as demonstrated by the performance of DeepSeek-Coder-BC, equips the model with the ability to excel in LRPL contexts. By using the code-bridge as an intermediate step, our method enhances the quality of LRPL training data, making it more aligned with the model’s existing strengths in HRPLs. This combination, along with a carefully designed training strategy, allows the model to effectively adapt to LRPLs. The significant performance gains across all benchmarks highlight that Bridge-Coder successfully addresses the scarcity of NL-PL aligned data in LRPLs, resulting in more accurate and coherent code generation in these underrepresented languages.

\begin{table}[h]
\begin{minipage}[c]{0.5\textwidth}

    \centering
    \resizebox{1\textwidth}{!}
{
\setlength{\extrarowheight}{2pt} 
\begin{tabular}{c|ccc}
\Xhline{3\arrayrulewidth}  
\textbf{Training Methods} & \textbf{R} & \textbf{D} & \textbf{Avg}\\ 
\Xhline{3\arrayrulewidth}
Separate Alignment & 46.89 & 23.87 & 35.38 \\
Direct Alignment & 42.63 & 32.45 & 37.54 \\
Assist Alignment & 42.93 & 33.87 & 38.40 \\
Bridged Alignment & \textbf{49.11} & \textbf{35.51} & \textbf{42.31} \\
\Xhline{3\arrayrulewidth}  
\end{tabular}
}
\vspace{0.2cm}
\caption{Comparison of training methods.}
\label{exp: training}

\end{minipage}
\hspace{0.5cm}
\begin{minipage}[c]{0.45 \textwidth}
\centering

\resizebox{\textwidth}{!}{
\setlength{\extrarowheight}{2pt} 
\begin{tabular}{cc|ccc}
\Xhline{3\arrayrulewidth}  
\textbf{Synthesis} & \textbf{Transfer} & \textbf{R} & \textbf{D} & \textbf{Avg}\\
\Xhline{3\arrayrulewidth}
Code & Code & 29.56 & 26.61 & 28.08\\
Code & General & 32.22 & \underline{27.50} & 29.86\\
General & General & \textbf{37.53} & \textbf{28.16} & \textbf{32.85}\\
General & Code & \underline{34.23} & 25.90 & \underline{30.07}\\
\Xhline{3\arrayrulewidth}  
\end{tabular}

}
\vspace{0.2cm}
\caption{Comparison of code-specific models (Code) and general-purpose models (General) in different combinations of Code-Bridge Synthesis and Guided Code Transfer. \textbf{Bold} indicates the best result, and \underline{underline} indicates the second-best result.}
\label{exp: model_select_in_generation}
\end{minipage}
\end{table}

\subsection{Comparison of Training Methods}
\label{sec: training exp}
We compared various training methods to assess their effectiveness in aligning NL instructions with LRPL outputs. As shown in Table \ref{exp: training}, \textit{Assist Alignment} alone performs worse because the model becomes overly reliant on the code-bridge and struggles to generalize to NL-only instructions. \textit{Direct Alignment} also underperforms, as the model is forced to bridge the NL-PL gap without any support, highlighting the importance of gradual learning.
Our \textit{Bridged Training} approach, which begins with \textit{Assist Alignment} and transitions to \textit{Direct Alignment}, consistently achieves the best results. To ensure the improvements weren’t solely due to the HRPL component of the code-bridge, we tested \textit{Separate Alignment}, which showed instability in D, confirming that combining the two phases of \textit{Bridged Training} leads to more robust and effective performance..


\subsection{Further Analysis}

\subsubsection{Code LLM Performs Better?}
\label{analysis: model select}

One might expect that code-specific models would perform best for generating code-related data, but as shown in Table \ref{exp: model_select_in_generation}, the combination of code models for both Synthesis and Transfer stages actually performs the worst. In contrast, general-purpose models consistently improve performance, with the best results coming from using general models for both stages. This can be attributed to the fact that Code-Bridge Synthesis primarily leverages the model's HRPL capability, which reduces the performance gap between code-specific and general models. However, in the Guided Code Transfer stage, in-context learning (ICL) becomes more critical, where general models seem to outperform code-specific ones.

\begin{table}[h]
\begin{minipage}[c]{0.45\textwidth}

    \centering
    \resizebox{1\textwidth}{!}
{
\setlength{\extrarowheight}{2pt} 
\begin{tabular}{c|cc|c}
\Xhline{3\arrayrulewidth}  
\textbf{Assist Format} & \textbf{R} & \textbf{D} & \textbf{Avg} \\ 
\Xhline{3\arrayrulewidth}
PL & 42.64 & 32.57 & 37.61 \\
NL & 40.89 & 30.72 & 35.81 \\
NL + PL & \textbf{44.71} & \textbf{35.51} & \textbf{40.11} \\
\Xhline{3\arrayrulewidth}  
\end{tabular}
}
\vspace{0.2cm}
\caption{Comparison of different assist formats in the phase of Assist Alignment.}
\label{exp: bridge}

\end{minipage}
\hspace{0.5cm}
\begin{minipage}[c]{0.42 \textwidth}
\centering

\resizebox{\textwidth}{!}{
\setlength{\extrarowheight}{2pt} 
\begin{tabular}{c|cc}
\Xhline{3\arrayrulewidth}  
\textbf{Code-Bridge HRPL} & \textbf{R} & \textbf{D} \\ 
\Xhline{3\arrayrulewidth}
Python & \textbf{47.61} & \textbf{33.87} \\
C$++$ & 44.29 & 30.62 \\
JAVA & 46.47 & 31.93 \\
\Xhline{3\arrayrulewidth}  
\end{tabular}

}
\vspace{0.2cm}
\caption{Further Analysis of different Programming Languages used in Bridge-Coder.}
\label{exp: code select}

\end{minipage}
\end{table}


\subsubsection{NL vs. PL: Which Matters More?}
\label{analysis: comments}
In the first phase of our Bridged Training, we explored whether using NL-formatted comments or PL-formatted code as part of the Assist Alignment yields better alignment. As shown in Table \ref{exp: bridge}, training with code (PL) alone outperforms comments (NL) alone. However, relying solely on code is still not the optimal approach. The combination of both NL and PL (code \& comments) leads to the best results, highlighting the complementary nature of NL and PL in bridging the NL-PL gap and improving overall performance. This also explains why, in our generation of the code-bridge, we emphasize the need for explanations in the form of NL comments to assist and enhance the code output.


\subsubsection{Different HRPLs as Code-Bridge}
\label{analysis: PLs as code-bridge}
The results in Table \ref{exp: code select} demonstrate that Python outperforms C++ and Java as the code-bridge programming language. This is likely due to Python's prevalence in the training data, which enables the model to generate more accurate and effective code-bridges. Python's extensive library ecosystem for tasks like data science and automation also provides more tools for generating robust code. Additionally, Python's simplicity and readability contribute to better alignment with natural language instructions, facilitating a smoother NL-PL transition. In contrast, C++ and Java's more complex syntax and explicit logic make them less effective for generating efficient code-bridges in this context.

\begin{wrapfigure}[9]{r}{0.45\textwidth}
\begin{minipage}{.45\textwidth}
\vspace{-12mm}
    \includegraphics[width=0.9\columnwidth]{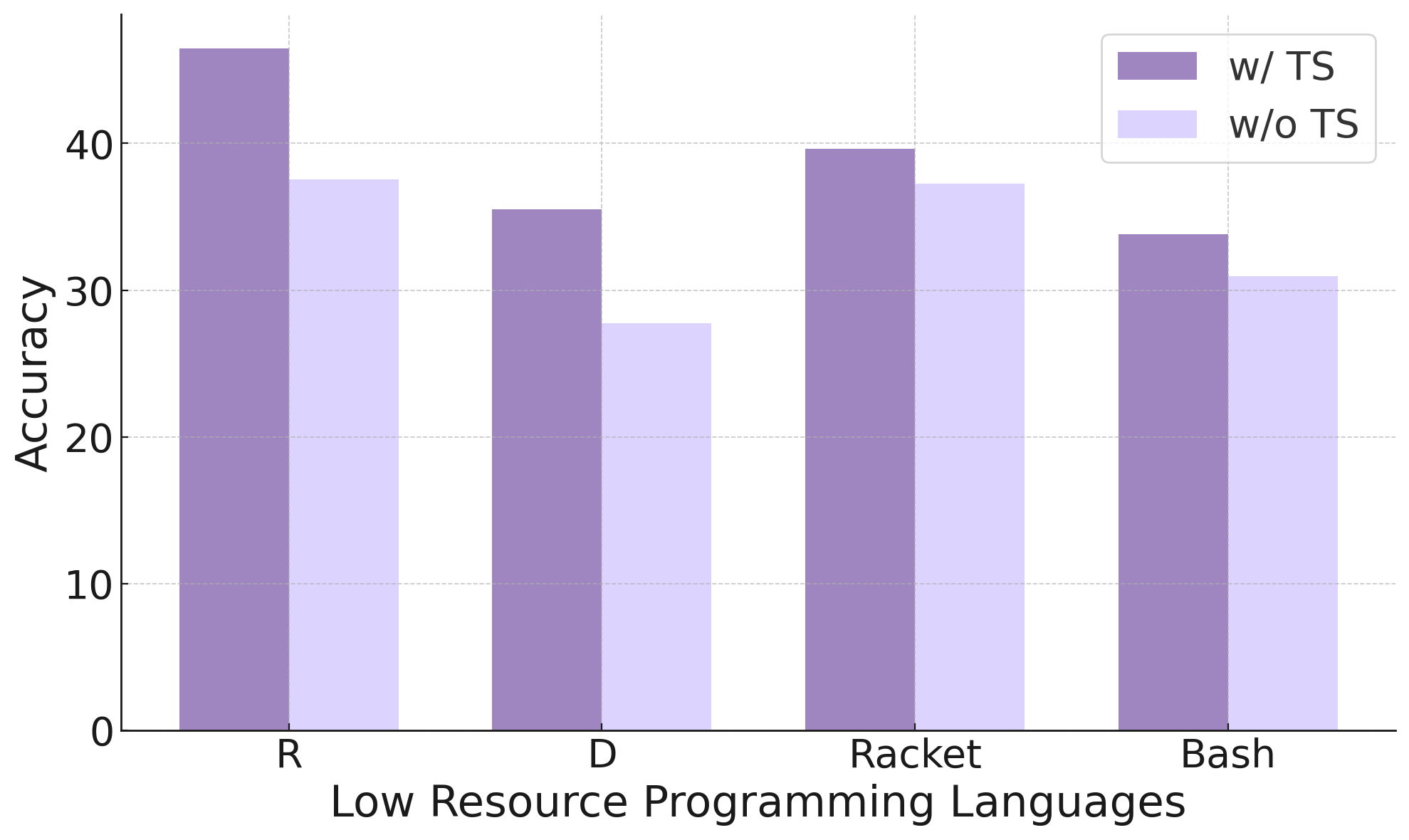}
    \caption{Ablation of Task Screening.}
    \label{exp: screening}
\end{minipage}
\end{wrapfigure}



\subsubsection{Ablation of Task Screening}
\label{analysis: task screening}
Figure \ref{exp: screening} highlights the importance of task screening. While the dataset without screening includes more tasks, the performance on unanswerable tasks is poor. With Task Screening (w/ TS), accuracy improves significantly across all LRPLs (R, D, Racket, Bash). This demonstrates that filtering out tasks beyond the model's capability leads to better results and validates the effectiveness of using LLMs' general reasoning for task screening.

\section{Conclusion}

This paper tackles the challenge of generating high-quality programs in low-resource languages using LLMs. By leveraging LLMs' instruction-following abilities and expertise in high-resource programming languages, we create a new, high-quality dataset for low-resource languages. We also propose a progressive training pipeline to maximize the use of these samples, improving instruction-following performance in low-resource languages. Experimental results show our methods significantly outperform the baseline.

\section*{Limitations}
Despite strong instruction-following capabilities, our work remains confined to repository-level text-to-code generation, which involves long-context modeling and resolving lost-in-the-middle issues. Additionaly, future studies should address multi-round text-code challenges, requiring repeated interactions and more detailed instructions. 


\input{template.bbl}

\bibliographystyle{unsrtnat}
\bibliography{template}

\newpage

\appendix

\section{Appendix}

\subsection{Detailed Prompts}
\label{appendix: prompts}

This is a section in the appendix.

\begin{table*}[h!]
\centering
\begin{minipage}{1.0\textwidth}
\vspace{0mm}    
\centering
\begin{sectionbox}[]{Prompt for Task Screening}
    \centering
      \footnotesize
    \begin{tabular}{p{0.97\textwidth} c}
You are a highly knowledgeable assistant with expertise in evaluating whether a given problem can be solved using various programming languages. \\
\vspace{0.5em}
You should judge whether \textcolor{olive}{$<$Programming Language$>$} can be used to solve the problem below. \\
\vspace{0.5em}
You should always respond with either ``Yes" or ``No", followed by a concise explanation. Be concise and direct in your responses.\\
\vspace{0.5em}
Here is the task: \textcolor{blue!70}{$<$Task$>$}\\

    \end{tabular}
\end{sectionbox}
\vspace{-2mm}
\caption{The prompt for screening tasks that are unanswerable in \textcolor{olive}{Low-Resource Programming Language}.}
\label{prompt: task screening}
\end{minipage}
\end{table*}

\begin{table*}[h!]
\centering
\begin{minipage}{1.0\textwidth}
\vspace{0mm}    
\centering
\begin{sectionbox}[]{Prompt for Bridge-Assisted Generation}
    \centering
      \footnotesize
    \begin{tabular}{p{0.97\textwidth} c}
You are a highly knowledgeable assistant with expertise in generating solutions across multiple programming languages, providing detailed explanations for each step. \\
\vspace{0.5em}
Help me use \textcolor{orange}{$<$Programming Language$>$} to solve the problem below. In your response, you need to provide \textbf{detailed comments} to explain the key steps and the reasoning process, rather than only responding the solution.\\
\vspace{0.5em}
Here is the task: \textcolor{blue!70}{$<$Task$>$}\\
    \end{tabular}
\end{sectionbox}
\vspace{-2mm}
\caption{The prompt for synthesizing code-bridge in \textcolor{orange}{High-Resource Programming Language}.}
\label{prompt: bridge synthesis}
\end{minipage}
\end{table*}

\begin{table*}[h!]
\centering
\begin{minipage}{1.0\textwidth}
\vspace{0mm}    
\centering
\begin{sectionbox}[]{Prompt for Guided Code Transfer}
    \centering
      \footnotesize
    \begin{tabular}{p{0.97\textwidth} c}
You are a highly knowledgeable assistant that specializes in problem-solving across various programming languages. \\
\vspace{0.5em}
Help me use \textcolor{olive}{$<$Programming Language$>$} to solve the problem below. \\
\vspace{0.5em}
Here is the task: \textcolor{blue!70}{$<$Task$>$}\\
\vspace{0.5em}
To help you better solve this task, you can refer to this solution in \textcolor{orange}{$<$Programming Language$>$}: \textcolor{orange}{$<$Code-Bridge$>$}\\

    \end{tabular}
\end{sectionbox}
\vspace{-2mm}
\caption{The prompt for generating answers in \textcolor{olive}{Low-Resource Programming Language}. \textcolor{orange}{$<$Code-Bridge$>$} is the answer in a \textcolor{orange}{High-Resource Programming Language}.}
\label{prompt: bridge synthesis}
\end{minipage}
\end{table*}

\subsection{Low-Resource Programming Languages}
\label{appendix: lrpl}

\begin{itemize}
    \item \textbf{R}: A programming language widely used for statistical computing, data analysis, and visualization. It is highly popular in academia, research, and data science due to its extensive libraries and tools for handling complex data.

    \item \textbf{D}: A systems programming language designed for high performance and productivity. It combines the power of C and C++ with more modern features, making it ideal for applications that require efficiency and low-level system access, while maintaining a developer-friendly syntax.

    \item \textbf{Racket}: A functional programming language that excels in language creation and experimentation. It is commonly used in academic settings and research for developing new programming languages, as well as for teaching concepts in computer science and functional programming.

    \item \textbf{Bash}: A Unix shell and command language widely used for scripting and automation tasks in system administration, software development, and task automation. Bash scripts are frequently used for managing servers, executing repetitive tasks, and automating workflows in Linux environments.
\end{itemize}

\section{Implementation Details}
For optimization, we used the Adafactor~\citep{shazeer2018adafactor} optimizer with a learning rate of \(5 \times 10^{-5}\). The model was trained for 2 epochs with a warm-up of 15 steps. The batch size was set to 512.
To improve efficiency, we employed Flash Attention~\citep{dao2022flashattention} and used the bf16  precision for faster and more memory-efficient training.

\end{document}

%% file: template.bbl
\providecommand{\CNFX}[1]{{\em{\textrm{(#1)}}}}